\documentclass[nohyperref]{article}

\usepackage[T1]{fontenc}

\usepackage{microtype}
\usepackage{graphicx}
\usepackage{subfigure}
\usepackage{booktabs} 

\usepackage{hyperref}
\hypersetup{
	colorlinks,
	citecolor=red,
	filecolor=black,
	linkcolor=blue,
	urlcolor=blue,
	backref
}

\usepackage[accepted]{icml2022}

\usepackage{amsmath}
\usepackage{amssymb}
\usepackage{mathtools}
\usepackage{amsthm}

\usepackage[capitalize,noabbrev]{cleveref}

\usepackage[ruled]{algorithm2e}
\usepackage{algpseudocode}
\usepackage{graphicx}
\usepackage{gensymb}
\usepackage{multirow}

\usepackage{pifont}

\newcommand{\cmark}{\ding{51}}%
\newcommand{\xmark}{\ding{55}}%

\icmltitlerunning{Continual learning on 3D point clouds with random compressed rehearsal}

\newcommand{\cf}{catastrophic forgetting}
\newcommand{\model}{RCR}
\newcommand{\modellong}{Random Compression Rehearsal}

\begin{document}
\twocolumn[
\icmltitle{Continual learning on 3D point clouds with random compressed rehearsal}

\icmlsetsymbol{equal}{*}

\begin{icmlauthorlist}
\icmlauthor{Maciej Zamorski}{1}
\icmlauthor{Michał Stypułkowski}{2}
\icmlauthor{Konrad Karanowski}{1}
\icmlauthor{Tomasz Trzciński}{3,4,5}
\icmlauthor{Maciej Zięba}{1,4}
\end{icmlauthorlist}

\icmlaffiliation{1}{Wroclaw Univeristy of Science and Technology, Wroclaw, Poland}
\icmlaffiliation{2}{University of Wroclaw, Wroclaw, Poland}
\icmlaffiliation{3}{Warsaw University of Technology, Warsaw, Poland}
\icmlaffiliation{4}{Tooploox, Wroclaw, Poland}
\icmlaffiliation{5}{Jaigellonian University of Cracow, Cracow, Poland}

\icmlcorrespondingauthor{Maciej Zamorski}{maciej.zamorski@pwr.edu.pl}

\icmlkeywords{Continual Learning, Point Cloud, Deep Learning, Data Compression}

\vskip 0.3in
]



\printAffiliationsAndNotice{}  

\begin{abstract}
    Contemporary deep neural networks offer state-of-the-art results when applied to visual reasoning, e.g., in the context of 3D point cloud data.
	Point clouds are important datatype for precise modeling of three-dimensional environments, but effective processing of this type of data proves to be challenging.
	In the world of large, heavily-parameterized network architectures and continuously-streamed data, there is an increasing need for machine learning models that can be trained on additional data. Unfortunately, currently available models cannot fully leverage training on additional data without losing their past knowledge.
	Combating this phenomenon, called \textit{\cf{}}, is one of the main objectives of continual learning.
	Continual learning for deep neural networks has been an active field of research, primarily in 2D computer vision, natural language processing, reinforcement learning, and robotics.
	However, in 3D computer vision, there are hardly any continual learning solutions specifically designed to take advantage of point cloud structure.
	This work proposes a novel neural network architecture capable of continual learning on 3D point cloud data.
	We utilize point cloud structure properties for preserving a heavily compressed set of past data.
	By using rehearsal and reconstruction as regularization methods of the learning process, our approach achieves a significant decrease of \cf{} compared to the existing solutions on several most popular point cloud datasets considering two continual learning settings:  when a task is known beforehand, and in the challenging scenario of when task information is unknown to the model.
\end{abstract}
	
	\section{Introduction}
	\label{Introduction}
	In recent years the application of deep learning in computer vision has been steadily rising due to its versatility and performance~\cite{he2016deep,dosovitskiy2020image,qi2017pointnet,qi2017pointnet++,wu2019pointconv}.
	However, due to the growing number of parameters of machine learning models and the large amount of data needed for training, continuous improvement of models' performance poses a challenge from the computational expensiveness standpoint.
	When re-training on a new task, the effectiveness of neural network performance on the past tasks quickly degrade~\cite{delange2021continual,hsu2018re,masana2020class}.
	This phenomena is called \textit{catastrophic forgetting}.
	Finding an effective way to avoid it is a main object  of research in the field of continual learning~\cite{mai2021online,masana2020class,belouadah2020comprehensive,delange2021continual,parisi2019continual}, with most of the current research focusing on image~\cite{rusu2016progressive,li2017learning,kirkpatrick2017overcoming,shin2017continual,zenke2017continual,kirichenko2021task} and video data~\cite{doshi2020continual,doshi2022rethinking}.
	The further natural direction, being an application to the domain of 3D computer vision, is only just starting to be explored~\cite{chowdhury2021learning,dong2021i3Dol}.
	
	Nevertheless, the existing approaches require either one or a combination of 1) training one or many generative models to produce replayed samples~\cite{shin2017continual}, 2) keeping a snapshot of all parameters of the model to produce soft-labels for past data, therefore, restricting the generalization ability~\cite{li2017learning,chowdhury2021learning}, or 3) selecting only a small subset of exemplars or creating the set of primitives based on past data~\cite{rebuffi2017icarl,dong2021i3Dol}.
	These factors usually make them inadequate when dealing with a large number of tasks as well as computationally expensive to train.
	
	In this work we introduce \textit{\modellong{} (\model{})} -- a~novel approach to continual learning for classification of 3D point clouds.
	The \model{} makes use of basic autoencoder-like architecture and shares the latent representation of the input point cloud with the classification component.
	During the training, the reconstruction loss serves as an additional regularization term. 
	Moreover, we introduce a new rehearsal technique to compress point clouds from previous tasks.
	By storing just a small number of points from the past training datasets, we are able to effectively mitigate the \cf{} without the need for training an additional generative model, keeping a snapshot for the inference of soft labels, or introducing several additional layers to our classifier.
	
	The advantages of the \modellong{} are as follows:
	\begin{itemize}
		\item Classification accuracy -- \model{} achieves lower forgetting for point cloud data compared to the existing approaches, even after many instances of re-training
		\item Small computational cost -- Our model consists only of a feature extractor backbone with classification and decoder modules, each composed of a small number of trainable parameters. 
		Moreover, unlike the methods that rely on replay data from generative modeling, our work does not require training a generative model and utilizes data reconstruction only as a regularization technique for the backbone network.
		\item Storage efficiency -- For effective catastrophic forgetting mitigation, the \model{} requires storing a tiny portion of original training data for rehearsal purposes.
	\end{itemize}	

	In summary, the contribution presented in this paper is a new architecture, called \modellong{} along with an extensive evaluation conducted on 2D and 3D point cloud benchmark datasets.           
	
	The work is structured as follows: 
	\Cref{sec:related} introduces relevant concepts and previous work related to our paper,
	In \cref{sec:preliminaries} we outline the necessary background on point clouds.
	\Cref{sec:method} describes the \modellong{} in detail.
	In \cref{sec:experiments}, we present the empirical evaluation of our model and discuss the results of the experiments as well as an ablation study.
	We offer a summary of the work in \cref{sec:conclusion}.

	\section{Related work}
	\label{sec:related}
	This section reviews the relevant work in continual learning, focusing on its applications for 3D data.
	
	\subsection{Continual learning}
	\label{sec:related_cl}
	Several strategies to mitigate catastrophic forgetting have been proposed in recent years.
	Those strategies can be broadly categorized into several groups: architecture growing, training with additional regularization, saving subsets of data as a memory, and replay/rehearsal.
	Continual learning methods often use more than one of those strategies, making strict categorization difficult.
	
	In \cite{rusu2016progressive} the authors propose an approach called \textit{progressive networks} that employs progressively adding new layers and connections with each new task. 
	Thanks to these connections, the method may utilize past knowledge for solving future tasks.
	However, adding new layers makes the network grow linearly with each added task.
	
	Learning Without Forgetting~\cite{li2017learning} also assumes a network-growing approach.
	While it removes the need to store past data before the re-training, the model snapshot is stored instead.
	This snapshot is then used to obtain the last layer's features for new samples, thus creating \textit{soft labels}. 
	The model weights are regularized by matching the output values to the soft labels in the previous task range.
	
	The authors of Elastic Weight Consolidation \cite{kirkpatrick2017overcoming} depart from the idea of growing networks and assume that current deep neural networks are heavily overparametrized. 
	The focus is on embedding the new knowledge using those excessive parameters without changing the crucial regions for past data.
	It is done by estimating the importance of each parameter for solving already learned tasks and penalizing the gradient change of those parameters.	
	However, importance calculation is needed to be done separately for each of the previous tasks, and as such, the necessary computational cost makes it challenging to scale.
	
	Deep Generative Replay \cite{shin2017continual} adds a generative model that learns the distribution of previous data and replays it when optimizing for the next task. 
	This approach was the first to show good results in a challenging problem in which the information about the task is not provided during classification and needs to be inferred or skipped.
	The drawback of such a method is the necessity of training one or more generative models that approximate the past data distribution.
	The training of such models is often computationally demanding and requires a large number of samples to sufficiently cover the distributions mentioned above.
	This disadvantage can be especially seen when dealing with many tasks.
	With the influx of new data, the quality of generated samples belonging to the previous classes starts degrading, shifting the replayed distribution away from the original one.
	
	In iCARL~\cite{rebuffi2017icarl}, the authors propose the method that uses the \textit{memory buffer}, a fixed-size set of specially selected samples from the previous data that is updated after each new task.
	At the evaluation time, the samples from the buffer are used to classify new data with a softmax classifier or simple classification methods, such as the nearest neighbor.
	However, the updates to the memory buffer are done at the end of the training for each task.
	During the update procedure, the algorithm keeps the optimal samples and discards the rest. 
	This process depends on data and is therefore sensitive to ordering.
	
	In \cite{von2019continual}, the authors leverage hypernetworks~\cite{ha2016hypernetworks}, the model based on two modules -- trainer and target networks.
	During the optimization routine, the trainer network learns to generate the weights for the target network, which are conditioned on the characteristics of the input sample (e.g., task number).
	This approach allows obtaining a dedicated target function for each of the tasks presented to the trainer.
	However, hypernetwork training is challenging as it requires careful parameter tuning to obtain stable convergence, which renders the proposed method difficult to apply.

	\subsection{Continual Learning on Point Clouds}
	The usage of continual learning has also been emerging in application to 3D data.
	
	In~\cite{chowdhury2021learning} authors present an extension of LwF~\cite{li2017learning} architecture to work on point clouds.
	They achieve this by generating output vectors that, in addition to the input shape, are based on semantic representation embeddings of introduced classes. 
	The trained network is then penalized for the distribution change between current output vectors and those obtained from the snapshot. 
	
	A different approach, presented in I3DOL~\cite{dong2021i3Dol}, employs an attention mechanism to construct geometric primitives.
	Catastrophic forgetting is combated by storing a discriminative set of exemplars.
	During test time, they are compared to regions of an input data sample proposed by attention.
	The drawback of this solution is a highly complex architecture and difficulty in selecting the number of exemplars.
	Another risk is choosing structures that are potentially similar to those in future classes, which can lead to misclassification of new data or forgetting essential knowledge from the past.
	
	\section{Preliminaries}
	\label{sec:preliminaries}
	\subsection{Autoencoders}
	The \textit{(bottleneck) autoencoders} are the class of models consisting of encoding and decoding modules that are useful for learning a low-dimensional representation of data.
	Such representations, called \textit{feature vectors} are often used as data embeddings for downstream tasks.
	Training autoencoders is usually performed by minimizing the reconstruction error between the original input and the sample decoded from the features, which we can write down as
	\begin{equation}
		\mathcal{L} = d(x, D(E(x))),
	\end{equation}
	where $\mathcal{L}$ denotes the training loss, $E,D$ are the encoding and decoding modules, $d(\cdot, \cdot)$ is a distance function and $x$ is an input to the model.
	
	\subsection{Point clouds}
	A \textit{point cloud} is a data type represented by a set of points in a given multi-dimensional space.
	It is one of the most common ways to store and represent 3D data gathered by the scanning devices such as LIDAR or multi-camera setups.
	We can define an $N$-dimensional point cloud $X$ as a set of $K$ real-valued points, such as 
	\begin{equation}
		X = \{ \mathbf{x}_1, \mathbf{x}_2, \ldots, \mathbf{x}_K\},
	\end{equation}
	where each point $\mathbf{x}_k$ is represented by a vector of size $N$, that we can write as
	\begin{equation}
		\mathbf{x}_k = [ x_{k,1}, \ldots, x_{k, N} ]^{\mathrm{T}}
	\end{equation}
	
	However, due to the unordered nature of sets, processing point cloud data may be computationally challenging.
	This difficulty arises because the point clouds are permutation invariant, i.e., a sample consisting of $K$ points may be stored in any of $K!$ different orderings that ultimately represent the same object.
	It is especially problematic when there is a need to compare generated point cloud data, as is the case when training an autoencoder.
	The autoencoder training procedure usually requires a function to calculate the distance between original and reconstructed data. 
	Because of the permutation invariance, when using the distance function used in the image domain (e.g., $L^2$ distance), the resulting distance between two orderings that ultimately result in the same object may be greater than zero.
	Therefore, to represent the distance between two point clouds, one needs to use the metric calculating the distance between sets.
	The most common are the Chamfer Distance (CD) and the Earth Mover's (also called Wasserstein) Distance~\cite{rubner2000earth} (EMD).
	
	Chamfer Distance -- \cref{eq:cd} -- between sets $X_1$ and $X_2$ is defined as a sum of the shortest distances from each point in $X_1$ to it's closest point in $X_2$ and analogically, from each point in $X_2$ to its closest point in $X_1$.
	\begin{equation}
		\label{eq:cd}
		\begin{split}
		CD(X_1,X_2) =& \sum_{\mathbf{x}_1 \in X_1} \min_{\mathbf{x}_2\in X_2} \lVert \mathbf{x}_1-\mathbf{x}_2 \rVert_2^2 \\
		             &+ \sum_{\mathbf{x}_2 \in X_2} \min_{\mathbf{x}_1\in X_1} \lVert \mathbf{x}_1-\mathbf{x}_2 \rVert_2^2
		\end{split}.
	\end{equation}
	On the other hand, Earth Mover's Distance -- \cref{eq:emd} -- solves the optimal transport problem~\cite{kolouri2017optimal} between point clouds $X_1$ and $X_2$, i.e., finds a bijection between input sets that minimizes the total distance between matched points.
	It can be written as
	\begin{equation}
		\label{eq:emd}
		EMD(X_1,X_2) =\min_{\phi:X_1\to X_2} \sum_{\mathbf{x} \in X_1} \frac{\left\lVert \mathbf{x} - \phi(\mathbf{x}) \right\rVert_2^2}{2} ,
	\end{equation}
	where $\phi$ is a bijection.
	However, due to the vast search space of possible pairings, calculating the exact value of an EMD is computationally infeasible~\cite{villani2009optimal}.
	While using an approximate dual problem of EMD may be possible, implementing it correctly and efficiently is non-trivial.
	Therefore, in this work, we opt for using a Chamfer Distance as a much more lightweight and efficient method of calculating the point cloud difference.
	
	\section{Method}
	\label{sec:method}
	
	In this section, we offer a precise formulation of a problem we are aiming to solve and present our \modellong{} model.
	
	\begin{figure*}[ht!]
		\centering
		\includegraphics[width=0.8\textwidth]{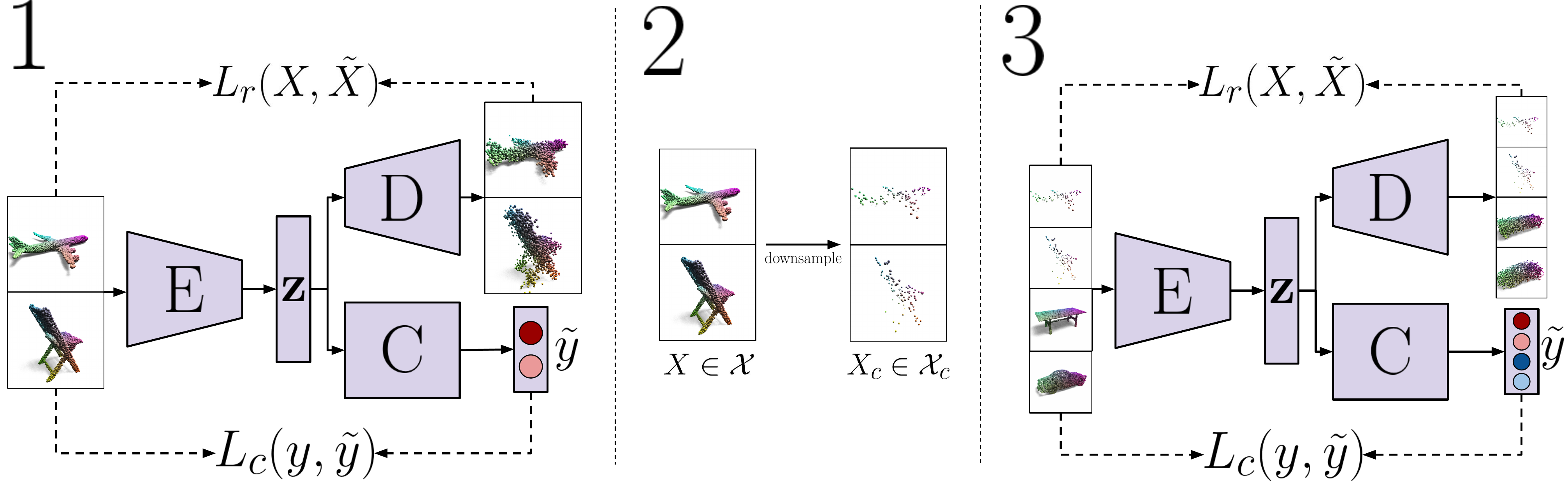}
		\caption{
			The training procedure of \modellong{}. 
			1) Train the model to reconstruct an input shape $X$ using a decoder $D$ and predict its class $y$ using a classifier $C$ simultaneously based on shared features $\mathbf{z}$ extracted by PointNet backbone $E$.
			Optimize the weights based on reconstruction loss $L_r$ and classification loss $L_c$ (see \cref{eq:loss}) criteria for outputs $\tilde{X}$ and $\tilde{y}$ . 
			2) Randomly (with uniform probability) select a subset of each point cloud $X$ in the training dataset $\mathcal{D}$, creating the compressed dataset $\mathcal{D}_r = \left\{ X_c \subset X, y)\ \vert\ (X, y) \in \mathcal{D} \right\}$ for rehearsal. 
			3) During retraining, use an union of new and compressed datasets (i.e. $\mathcal{D}_{new} \cup \mathcal{D}_r$) and expand the last layer of the classifier network $C$ to accommodate for more classes.}
		\label{fig:arch}
	\end{figure*}
	
	\subsection{Problem formulation}
	Let's assume that a model $f$ was trained on a dataset $\mathcal{D}$ of $M$ point clouds consisting of $K$ $N$-dimensional points with corresponding class labels, i.e.
	\begin{equation}
		\mathcal{D} = \{ (X_m, y_m ) \}_{m=1}^M,
	\end{equation}
	where $y_m$ is the corresponding class label for the point cloud $X_m$.  
	For a considered dataset, $y_{m} \in Y$, where $Y$ is the set of possible class labels.
	
	The goal is to re-train the model on a given new dataset $\mathcal{D}_{new}$, that contains samples belonging to a different class domain $Y_{new}$, i.e.
	\begin{equation}
		Y \cap Y_{new} = \varnothing,
	\end{equation}
	while keeping to the minimum the \cf{} effect, i.e. the degrading of classification performance on samples from the original dataset $\mathcal{D}$.
	
	In this work, we will refer to complete training on data with newly introduced class labels as \textit{a task}. 
	For example, the model trained first on a dataset consisting of point clouds with labels "0" and "1" and then adapted to point clouds representing classes "2" and "3" was trained on two tasks. 
	During training, we aim to correctly classify all the samples up to the most recent task without the need to store all of the previously seen data.
	We will evaluate work on two continual learning scenarios, as described in~\cite{van2019three} -- \textit{task incremental learning} (\textit{Task-IL}, i.e. solve task, knowing which one it is beforehand) and \textit{class incremental learning} (\textit{Class-IL}, i.e. solve the task after inferring it).
	
	\subsection{\modellong{}}
	\label{sec:rcr}
	In this section, we present the proposed architecture of \modellong{}. 
	We aim to create a model that is able to capture general and high-level, as well as discriminative and low-level features of 3D point cloud shapes.
	While detailed features are fundamental to training any classifier, extracting the all-around representation provides twofold benefits:
	\begin{itemize}
		\item regularization method -- the application of autoencoder architecture that shares the latent space and uses reconstruction loss, prevents classifier from overfitting,
		\item future-proofing -- mitigates \cf{} by conditioning classifier on features that may be useful in future tasks.
	\end{itemize}
	We employ joint discriminative modeling and representation learning to ensure the embedding of both types of features.
	Discriminative modeling allows us to learn specific, low-level features, while the representation learning approach focuses more on the broad properties of the data samples.
	Therefore, our architecture consists of three neural networks: the common encoder backbone $E$, the classifier $C$, and the decoder $D$ as presented in \cref{fig:arch}.
	In this case, the encoder extracts latent representations of the given 3D shapes.
	On the other hand, the classifier and the decoder are being optimized for the purpose of classification and reconstruction.
	Thus they tug the focus of an encoder between discriminative and more general features.
	
	In addition to the aforementioned regularizations on an encoder $E$, we have also employed the rehearsing method to prevent the occurrence of catastrophic forgetting.
	Existing methods, primarily designed for images, implement rehearsing by supplying the training process with a portion of previous data~\cite{rebuffi2017icarl,yoon2021online}.
	Contrary to that approach, we take advantage of the fact that point clouds can be easily compressed.
	Therefore, we opt for random undersampling to store the entire past training data to be used for re-training the model for future tasks (see \cref{sec:ablation} for justification of compression method selection).
	
	\subsection{Model training and inference}
	In this section, we describe the training procedure of \modellong{}.
	
	Our continual learning procedure is done as follows.
	We begin by randomly initializing the parameters of all modules (encoder $E$, decoder $D$, and classifier $C$).
	Next, $E$, $D$ and $C$ are trained on the first task's data $D_1$ until convergence of loss function $L$ (as given in \cref{eq:loss}) 
	\begin{equation}
		\label{eq:loss}
		\begin{split}
			L(X, y, \tilde{X}, \tilde{y}) &= L_c(y, \tilde{y}) + L_r(X, \tilde{X})  \\
			& = \text{CrossEntropy}(y, \tilde{y}) \\
			& \quad + \text{ChamferDistance}(X, \tilde{X}),
		\end{split}
	\end{equation}
	where ChamferDistance as in eq. \eqref{eq:cd}, $X, y$ refer to original point cloud and its label, while $\tilde{X}, \tilde{y}$ refer to ones returned by $D$ and $C$ respectively.
	We then compress $D_1$ by replacing each point cloud in the dataset with a randomly subsampled, much smaller subset of itself.
	This compressed set creates our rehearsal dataset $D_r$ that will prevent catastrophic forgetting in the following tasks.
	For re-training the model on the next task's data, we again train it until convergence of the loss function $L$ on joint dataset $D_2 \cup D_r$, after which we compress the new data (in this case $D_2$) in the aforementioned fashion and add it to $D_r$ set for future rehearsal.
	The presented method is then repeated for each consecutive task.
	We present the described routine in pseudocode in algorithm~\labelcref{al:procedure}.

	After training, we discard the decoder module and use only the classifier $C$ with the encoder backbone $E$.
	For classification purposes, we perform feature extraction and classification in a fashion similar to the one introduced in PointNet\cite{qi2017pointnet}.
	
	\begin{algorithm}[ht!] \label{al:procedure}
		\caption{
    		\modellong{} training procedure for $T$ tasks. By $A \subset_K B$ we denote a point cloud A that is a random sub-sampling consisting of K points form point cloud B.
		}
		\SetKwInOut{Input}{In}
		\SetKwInOut{Output}{Out}
    	\DontPrintSemicolon
    	\small
		\Input{%
			$t \in \{1, \ldots, T\}$ -- current task number,
			$\mathcal{D}_t = \{ (X_1^t, y_1^t), \ldots, (X_N^t, y_N^t) \}$ -- training samples with corresponding class labels for the task $t$
		}
		\Output{%
        	$\theta_E^*, \theta_D^*, \theta_C^*$ -- optimal parameters for Encoder, Decoder and Classifier,
        	$\mathcal{D}_r$ -- Dataset of compressed past data used for rehearsal
		}
		
    	$\theta_E, \theta_D, \theta_C \leftarrow $ Random initialization\;
    	$\mathcal{D}_r \leftarrow \varnothing$ \tcp*{Dataset of rehearsal data}
		
		\For{$t \leftarrow 1$ \KwTo $T$}{
			\While{convergence not reached}{ 
				$X, y \leftarrow $ sample minibatch from $\mathcal{D}_t \cup \mathcal{D}_r$\;
				$z \leftarrow Encoder(X)$\;
				$\tilde{X} \leftarrow Decoder(z)$\;
				$\tilde{y} \leftarrow Classifier(z)$\;
				
				$L \leftarrow L_r(X, \tilde{X}) + L_c(y, \tilde{y})$ \tcp*{loss function as in \cref{eq:loss}}
				$\theta_E, \theta_D, \theta_C \leftarrow optimizer(\theta_E, \theta_D, \theta_C, \nabla L)$ \;
			}
			\tcp{Compress training data $\mathcal{D}_t$ from current task and add it to rehearsal set $\mathcal{D}_r$}
			$\mathcal{D}_r \leftarrow \mathcal{D}_r \cup \{ (\tilde{X}_t \subset_K X_t, y_t)\ \lvert\ (X_t, y_t) \in \mathcal{D}_t\}$\;
		}
	\end{algorithm}
	
	\section{Experiments}
	\label{sec:experiments}
	Here we discuss the experiments conducted to assess the quality of our approach.
	First,in \cref{sec:implementation} we describe in detail the implementation of \modellong{}.
	Then, in \cref{sec:results} we present the evaluation results of \model{} and compare its classification accuracy against the other popular methods from the literature.
	We examine the influence of different compression strategies and the usage of the reconstruction module on the model's quality in the ablation study in \cref{sec:ablation}.
	
	\begin{figure*}[htb!]
		\centering
		\includegraphics[width=0.8\textwidth]{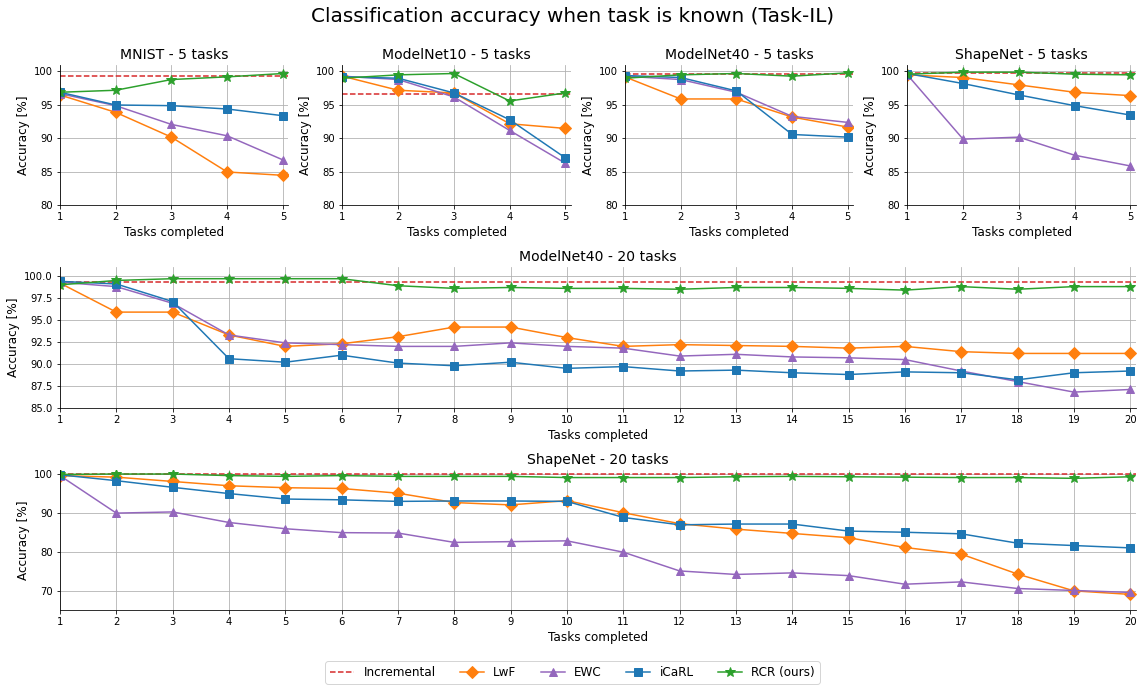}
		\caption{
			Average accuracy after $t$ tasks completed.  
			The task is \textit{known} during the inference.
			All models have a length of the feature vector set to 2048.
			The \model{} saves 100 points of each point cloud for rehearsal.
			We set the size of the memory buffer for iCaRL proportionally to the compression ratio of \model{} (i.e., 5\% of the total number of samples).
		}
		\label{fig:cl1}
	\end{figure*}
	
	\subsection{Implementation details}
	\label{sec:implementation}
	We implement all modules of \modellong{} as convolutional and feed-forward neural networks using the PyTorch library and train them on two NVIDIA GeForce RTX 2080Ti cards.
	
	Encoding backbone and classifier module with their hyperparameters were chosen as described in the original PointNet implementation~\cite{qi2017pointnet}.
	For the encoder, we use five 1-dimensional convolutional layers (consisting of $64-64-64-128-\lvert \mathbf{z} \rvert$ neurons, where $\lvert \mathbf{z} \rvert$ is the size of the latent dimension) with Batch Normalization, ReLU activations between them and the $\max(\cdot)$ aggregation function after the last layer.
	
	For the decoder, we use five fully-connected layers (consisting of $64-128-512-1024- M*N$ neurons, where $M$ is the fixed number of points in the point cloud and $N$ is the number of dimensions of the points in the dataset) with ReLU activations between them.
	
	For classifier we use 3 fully-connected layers (consisting of $512-256-\lvert \mathbf{c} \rvert$ neurons, where $\lvert \mathbf{c} \rvert$ is the number of the data classes) with Dropout ($p = 0.3$) and Batch Normalization (momentum $ = 0.5$) between them.
	
	Following \cite{qi2017pointnet}, batch normalization momentum in the encoder and classifier was subjected to exponential decay by $0.5$ every 100000 samples seen during training, with a lower limit of $0.01$.
	
	The parameters were obtained using an Adam optimizer, with a learning rate set to $10^{-3}$ at the beginning of each training and exponentially decayed by $0.7$ every 100000 steps to the minimum of $10^{-5}$, $\beta$ values equal to $(0.9, 0.999)$, without weight decay or AMSGrad.
	
	Each task was trained for a maximum of 1000 epochs, with a 100 epoch limit of no improvement.
	
	Compression by subsampling is performed by choosing the point clouds of the original shape with the uniform probability on each point.
	We chose to reduce the size of the original point clouds to 100 points, which is less than $5\%$ of their original size of 2048.
	
	For each minibatch, starting from task 2, we impose a minibatch composing condition that allocates half the minibatch size to the new and the other half to the rehearsal data.
	
	\begin{figure*}[htb!]
		\centering
		\includegraphics[width=0.8\textwidth]{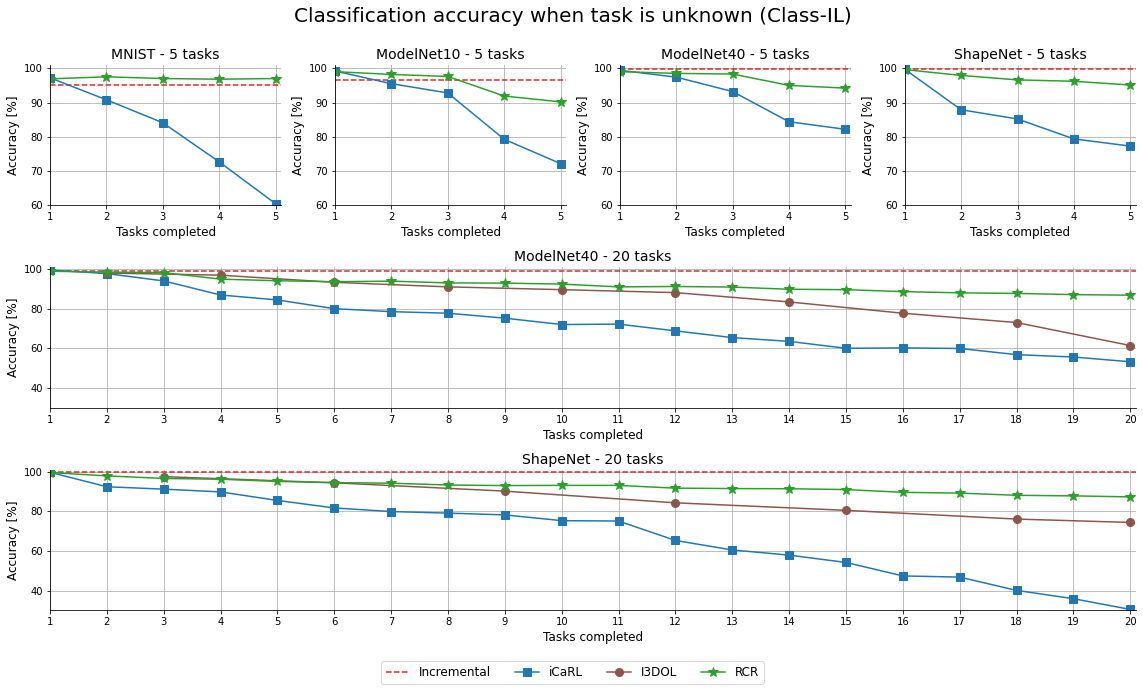}
		\caption{
			Average accuracy after $t$ tasks completed.  
			The task is \textit{unknown} during the inference.
			All models have a length of the feature vector set to 2048.
			The \model{} saves 100 points of each point cloud for rehearsal.
			We set the size of the memory buffer for iCaRL proportionally to the compression ratio of \model{} (i.e., 5\% of the total number of samples).
			The results for I3DOL as reported in~\cite{dong2021i3Dol} (note: I3DOL results are reported when training with 4 and 6 classes per task for ModelNet 40 and ShapeNet, respectively. 
			Error for ShapeNet trained on 40 classes was obtained by interpolating error values for 36 and 42 classes.).
		}
		\label{fig:cl3}
	\end{figure*}
	
	\subsection{Results}
	\label{sec:results}
	\begin{table*}[hbt!]
		\centering
		\caption{
		Final classification results after training for the specified number of tasks obtained by our model \model{} compared to baseline methods (supervised training, fine-tuning and incremental training) as well as leading approaches to continual learning adapted to point clouds data.
		Class-IL values for LwF and EWC are omitted since these models assume knowledge about the task.
		(note: I3DOL results are reported when training with 4 and 6 classes per task for ModelNet 40 and ShapeNet, respectively. 
		Error for ShapeNet trained on 40 classes was obtained by interpolating error values for 36 and 42 classes.)
		}
		\label{tab:results}
		\resizebox{\textwidth}{!}{%
            \begin{tabular}{@{}crrrrrrrrcrrrccrr@{}}
            \toprule
            \multicolumn{1}{c}{\multirow{2}{*}{Dataset}} & \multicolumn{1}{c}{\multirow{2}{*}{Tasks}} & \multicolumn{1}{c}{\multirow{2}{*}{Supervised}} & \multicolumn{2}{c}{Fine-tune} & \multicolumn{2}{c}{Incremental} & \multicolumn{2}{c}{LwF} & \multicolumn{2}{c}{EWC} & \multicolumn{2}{c}{iCaRL} & \multicolumn{2}{c}{I3DOL~\cite{dong2021i3Dol}} & \multicolumn{2}{c}{RCR} \\ \cmidrule(l){4-17} 
            \multicolumn{1}{c}{} & \multicolumn{1}{c}{} & \multicolumn{1}{c}{} & \multicolumn{1}{c}{Task-IL} & \multicolumn{1}{c}{Class-IL} & \multicolumn{1}{c}{Task-IL} & \multicolumn{1}{c}{Class-IL} & \multicolumn{1}{c}{Task-IL} & \multicolumn{1}{c}{Class-IL} & Task-IL & \multicolumn{1}{c}{Class-IL} & \multicolumn{1}{c}{Task-IL} & \multicolumn{1}{c}{Class-IL} & Task-IL & Class-IL & \multicolumn{1}{c}{Task-IL} & \multicolumn{1}{c}{Class-IL} \\ \midrule
            MNIST & 5 & 98.3 $\pm$ 0.1 & 71.2 $\pm$ 5.5 & 19.5 $\pm$ 0.3 & 99.3 $\pm$ 0.1 & 95.1 $\pm$ 0.4 & 84.5 $\pm$ 0.3 & - & 86.8 $\pm$ 0.1 & - & 93.4 $\pm$ 0.3 & 60.5 $\pm$ 1.8 & - & - & \textbf{99.7} $\pm$ 0.1 & \textbf{97.0} $\pm$ 0.1 \\ \midrule
            ModelNet10 & 5 & 92.1 $\pm$ 0.2 & 72.3 $\pm$ 1.1 & 15.0 $\pm$ 0.1 & 96.6 $\pm$ 0.4 & 90.4 $\pm$ 0.2 & 91.5 $\pm$ 0.4 & - & 86.3 $\pm$ 0.8 & - & 87.1 $\pm$ 0.2 & 72.2 $\pm$ 1.5 & - & - & \textbf{96.8} $\pm$ 0.2 & \textbf{90.2} $\pm$ 0.1 \\ \midrule
            \multirow{2}{*}{ModelNet40} & 5 & 97.6 $\pm$ 0.2 & 52.4 $\pm$ 1.1 & 20.0 $\pm$ 0.1 & 99.7 $\pm$ 0.1 & 96.7 $\pm$ 0.2 & 91.7 $\pm$ 0.4 & - & 92.4 $\pm$ 0.8 & - & 90.2 $\pm$ 0.4 & 82.2 $\pm$ 2.4 & - & - & \textbf{99.8} $\pm$ 0.1 & \textbf{94.2} $\pm$ 0.2 \\
             & 20 & 88.2 $\pm$ 0.8 & 56.4 $\pm$ 2.7 & 1.7 $\pm$ 0.1 & 99.3 $\pm$ 0.1 & 88.6 $\pm$ 0.5 & 91.2 $\pm$ 0.4 & - & 87.1 $\pm$ 1.6 & - & 89.2 $\pm$ 0.3 & 53.3 $\pm$ 0.9 & - & 61.5 & \textbf{99.0} $\pm$ 0.1 & \textbf{86.9} $\pm$ 0.2 \\ \midrule
            \multirow{2}{*}{ShapeNet} & 5 & 97.3 $\pm$ 0.1 & 85.3 $\pm$ 2.8 & 8.4 $\pm$ 0.1 & 99.8 $\pm$ 0.1 & 96.5 $\pm$ 0.1 & 96.4 $\pm$ 0.6 & - & 85.9 $\pm$ 0.9 & - & 93.5 $\pm$ 1.4 & 77.3 $\pm$ 1.0 & - & - & \textbf{99.5} $\pm$ 0.1 & \textbf{95.1} $\pm$ 0.3 \\
             & 20 & 90.8 $\pm$ 0.2 & 62.4 $\pm$ 4.4 & 0.6 $\pm$ 0.1 & 99.8 $\pm$ 0.1 & 91.5 $\pm$ 0.3 & 69.1 $\pm$ 0.3 & - & 69.6 $\pm$ 0.6 & - & 81.0 $\pm$ 0.7 & 30.5 $\pm$ 4.2 & - & 74.4 & \textbf{99.2} $\pm$ 0.1 & \textbf{87.3} $\pm$ 0.1 \\ \bottomrule
            \end{tabular}%
		}
	\end{table*}
	
	The experiments were conducted on three popular point cloud benchmark datasets: ModelNet 10, ModelNet 40~\cite{wu20153D}, and ShapeNet~\cite{chang2015shapenet}.
	Additionally, a point cloud version of the well-known MNIST dataset has been created to evaluate our approach on 2D point clouds.
	
	For all datasets, we sample 2048 points per point cloud.
	All input datasets were augmented as follows: random rotation ($\pm 15 \degree$ MNIST, $\pm 180 \degree$ around vertical axis for the rest), random noise ($\epsilon \sim \mathcal{N}(0, 0.02)$ capped at 0.05), random flip (vs XZ or YZ plane with 50\% chance).
	
	The experiments were run at the distance of 5 (all datasets) and 20 tasks (only ModelNet 40 and ShapeNet) with two class labels allocated for each task.
	We group the classes into tasks by ordering them by the number of point clouds associated with the class -- i.e., in the first task, there are two classes with the most samples; in the second task, the third and the fourth most, and so on.
	If each class was equally populous (as in MNIST), classes were assigned to tasks in the standard Split MNIST order, as presented in~\cite{van2019three}.
	
	We set a size of a latent representation in each model to 2048 features (i.e. $\lvert \mathbf{z} \rvert = 2048$ in \cref{sec:rcr}).
	For \modellong{} we uniformly subsample 100 points from each point cloud for rehearsal.
	For iCaRL, we set the size of the memory buffer proportionally to the amount of the data we keep when using the \model{}. 
	Therefore, when saving 100 points out of 2048 in a point cloud, we set the iCaRL buffer size to $\frac{100}{2048} \approx 5 \%$ of the total amount of the data samples in the training set.
	
	All experiments were scored over three independent runs and are presented as a mean with a standard deviation value.
	For brevity, we omit the confidence intervals in the task-by-task plots (\cref{fig:cl1} and \cref{fig:cl3}).
	
	In \cref{fig:cl1} and \cref{fig:cl3} we  present the results of \modellong{} vs baseline methods task-by-task for Task-IL and Class-IL scenarios respectively.
	For the Class-IL scenario, we only provide baseline results for iCaRL and I3DOL since LwF and EWC are not designed to work with an unknown task as presented in their original form.
	
	\Cref{tab:results} presents final results with the overall accuracy for all of the methods, including reference values for supervised learning (training with all classes from the start), incremental topline (rehearsal with all of the past data), and fine-tuning bottom line (fine-tuning to new data with each re-training), as well as the results for the baseline models from the literature and our method.
	As we can observe, \model{} is able to maintain a classification accuracy level similar to our top line.
	This behavior does not change when we train a model with a more significant number of tasks (i.e., 20 instead of 5), whereas other methods suffer from a significantly diminished quality of the accuracy of the predictions.
	
	\begin{table*}[htb!]
		\centering
		\caption{Ablation study -- classification accuracy after 5 tasks per dataset.}
		\label{tab:ablation_5}
		\resizebox{\textwidth}{!}{%
			\begin{tabular}{@{}ccc
					>{\columncolor[HTML]{EFEFEF}}r 
					>{\columncolor[HTML]{EFEFEF}}r 
					>{\columncolor[HTML]{FFFFFF}}r 
					>{\columncolor[HTML]{FFFFFF}}r 
					>{\columncolor[HTML]{EFEFEF}}r 
					>{\columncolor[HTML]{EFEFEF}}r 
					>{\columncolor[HTML]{FFFFFF}}r 
					>{\columncolor[HTML]{FFFFFF}}r @{}}
				\toprule
				& Task &
				&
				\multicolumn{2}{c}{\cellcolor[HTML]{EFEFEF}$z = 100$} &
				\multicolumn{2}{c}{\cellcolor[HTML]{FFFFFF}$z = 1024$} &
				\multicolumn{2}{c}{\cellcolor[HTML]{EFEFEF}$z = 2048 (100)$} &
				\multicolumn{2}{c}{\cellcolor[HTML]{FFFFFF}$z = 2048$} \\
				Dataset & known? &
				Reconstruction &
				\multicolumn{1}{c}{\cellcolor[HTML]{EFEFEF}Argmax} &
				\multicolumn{1}{c}{\cellcolor[HTML]{EFEFEF}Random} &
				\multicolumn{1}{c}{\cellcolor[HTML]{FFFFFF}Argmax} &
				\multicolumn{1}{c}{\cellcolor[HTML]{FFFFFF}Random} &
				\multicolumn{1}{c}{\cellcolor[HTML]{EFEFEF}Argmax} &
				\multicolumn{1}{c}{\cellcolor[HTML]{EFEFEF}Random} &
				\multicolumn{1}{c}{\cellcolor[HTML]{FFFFFF}Argmax} &
				\multicolumn{1}{c}{\cellcolor[HTML]{FFFFFF}Random} \\ \midrule
				& & \xmark &
				93.3 $\pm$ 0.3 &
				\textbf{97.3 $\pm$ 0.9} &
				95.8 $\pm$ 0.3 &
				96.5 $\pm$ 0.4 &
				95.2 $\pm$ 0.3 &
				96.3 $\pm$ 0.8 &
				93.0 $\pm$ 0.1 &
				96.7 $\pm$ 0.1 \\
				& \multirow{-2}{*}{\cmark} &
				\cmark &
				94.8 $\pm$ 0.5 &
				96.8 $\pm$ 0.2 &
				94.4 $\pm$ 0.4 &
				\textbf{96.7 $\pm$ 0.2} &
				95.0 $\pm$ 0.4 &
				\textbf{96.8 $\pm$ 0.2} &
				95.6 $\pm$ 0.2 &
				\textbf{96.9 $\pm$ 0.2} \\ \cmidrule{2-11}
				& &
				\xmark &
				48.5 $\pm$ 0.4 &
				80.7 $\pm$ 0.8 &
				67.8 $\pm$ 0.5 &
				\textbf{91.6 $\pm$ 0.6} &
				53.4 $\pm$ 0.4 &
				83.2 $\pm$ 1.3 &
				67.7 $\pm$ 0.1 &
				\textbf{91.7 $\pm$ 0.1} \\
				\multirow{-4}{*}{ModelNet10} & \multirow{-2}{*}{\xmark} &
				\cmark &
				65.2 $\pm$ 1.2 &
				\textbf{89.5 $\pm$ 0.3} &
				68.0 $\pm$ 0.6 &
				90.1 $\pm$ 0.2 &
				70.1 $\pm$ 0.6 &
				\textbf{90.2 $\pm$ 0.1} &
				83.5 $\pm$ 0.2 &
				89.1 $\pm$ 0.1 \\ \midrule
				
				& & \xmark & 96.2 $\pm$ 0.8 & \textbf{99.5 $\pm$ 0.7} & 94.8 $\pm$ 1.2 & 99.2 $\pm$ 0.6 & 98.1 $\pm$ 1.3 & 99.4 $\pm$ 0.3 & 99.6 $\pm$ 0.1 & 99.3 $\pm$ 0.1 \\
				& \multirow{-2}{*}{\cmark} & \cmark & \textbf{99.5$\pm$ 1.0} & 99.2 $\pm$ 0.2 & 99.5 $\pm$ 0.4 & \textbf{99.7 $\pm$ 0.1} & 99.3 $\pm$ 0.3 & \textbf{99.8 $\pm$ <0.1} & 99.6 $\pm$ 0.1 & \textbf{99.8 $\pm$ <0.1} \\ \cmidrule{2-11}
				& & \xmark & 59.2 $\pm$ 4.4 & 75.4 $\pm$ 1.5 & 73.6 $\pm$ 2.2 & \textbf{95.9 $\pm$ 1.0} & 72.2 $\pm$ 1.9 & 88.4 $\pm$ 0.7 & 76.8 $\pm$ 0.1 & \textbf{96.7 $\pm$ 0.1} \\
				\multirow{-4}{*}{ModelNet40} & \multirow{-2}{*}{\xmark} & \cmark & 80.6 $\pm$ 0.9 & \textbf{91.2 $\pm$ 0.2} & 92.2 $\pm$ 1.1 & 94.7 $\pm$ 0.6 & 90.1 $\pm$  0.4 & \textbf{94.2 $\pm$ 0.2} & 93.0 $\pm$ 0.1 & 95.5 $\pm$ 0.1 \\ \midrule
				
				& & \xmark & 95.2 $\pm$ 0.7 & 97.9 $\pm$ 0.5 & 96.9 $\pm$ 1.1 & \textbf{99.7 $\pm$ 0.1} & 97.2 $\pm$ 0.7 & 98.2 $\pm$ 0.2 & 97.4 $\pm$ 0.2 & \textbf{99.7 $\pm$ 0.1} \\
				& \multirow{-2}{*}{\cmark} & \cmark & 98.7$\pm$ 0.3 & \textbf{99.3 $\pm$ 0.2} & 99.0 $\pm$ <0.1 & 99.4 $\pm$ 0.2 & 99.0 $\pm$ 0.3 & \textbf{99.5 $\pm$ 0.1} & 98.6 $\pm$ 0.1 & \textbf{99.7 $\pm$ 0.1} \\ \cmidrule{2-11}
				& & \xmark & 41.9 $\pm$ 2.1 & 59.9 $\pm$ 0.7 & 42.2 $\pm$ 1.1 & \textbf{96.5 $\pm$ 0.2} & 42.8 $\pm$ 1.0 & \textbf{96.0 $\pm$ 0.2} & 43.1 $\pm$ 1.0 & \textbf{97.1 $\pm$ 0.1} \\
				\multirow{-4}{*}{ShapeNet} & \multirow{-2}{*}{\xmark} & \cmark & 81.3 $\pm$ 1.5 & \textbf{93.1 $\pm$ 0.5} & 88.6 $\pm$ 1.9 & 95.4 $\pm$ 0.4 & 83.2 $\pm$  0.8 & 95.1 $\pm$ 0.3 & 85.7 $\pm$ 1.2 & 95.9 $\pm$ 0.3 \\ \midrule
				
				& & \xmark & 97.6 $\pm$ 0.9 & 95.9 $\pm$ 1.0 & 97.2 $\pm$ 0.6 & 94.5 $\pm$ 2.0 & 97.4 $\pm$ 0.8 & 96.0 $\pm$ 0.4 & 97.8 $\pm$ 0.2 & 96.2 $\pm$ 0.6 \\
				& \multirow{-2}{*}{\cmark} & \cmark & 96.5 $\pm$ 1.0 & \textbf{98.4 $\pm$ 0.2} & 98.9 $\pm$ 0.2 & \textbf{99.7 $\pm$ 0.1} & 98.5 $\pm$ 0.5 & \textbf{99.7 $\pm$ 0.1} & 98.7 $\pm$ 0.1 & \textbf{99.7 $\pm$ 0.1} \\ \cmidrule{2-11}
				& & \xmark & 44.2 $\pm$ 7.3 & 27.2 $\pm$ 8.2 & 83.4 $\pm$ 2.1 & 43.3 $\pm$ 5.2 & 79.5 $\pm$ 0.5 & 68.8 $\pm$ 2.5 & 85.6 $\pm$ 0.9 & 87.0 $\pm$ 0.3 \\
				\multirow{-4}{*}{MNIST} & \multirow{-2}{*}{\xmark} & \cmark & 29.2 $\pm$ 9.5 & \textbf{58.1 $\pm$ 0.3} & 70.9 $\pm$ 3.2 & \textbf{97.7 $\pm$ 0.1} & 65.2 $\pm$  1.8 & \textbf{97.0 $\pm$ 0.1} & 67.8 $\pm$ 1.5 & \textbf{97.5 $\pm$ 0.1} \\ \bottomrule
			\end{tabular}%
		}
	\end{table*}
	
	\subsection{Ablation study}
	\label{sec:ablation}
	In this section, we present an ablation study to examine the impact of:
	\begin{itemize}
		\item the usage of reconstruction module,
		\item the size of the feature vector $\mathbf{z}$, 
		\item the method for picking the points for rehearsal,
		\item the number of points saved for rehearsal,
		
	\end{itemize}
	on the final classification accuracy in Task-IL and Class-IL scenarios.
	
	We tried several methods to make the encoder focus as much on the overall shape of the given samples as well as on the most discriminative features for the current task.
	One such method was adding another neural network -- decoder, to produce reconstructions of the input based solely on the latent feature vector.
	The purpose of the decoder was to regularize the latent space and embed the high-level features.
	Because the training of the decoder is done simultaneously with the rest of the modules, we can train our model end-to-end without the need for alternating training between solver and generator (as is the case in~\cite{shin2017continual}).
	
	The size of the bottleneck is usually one of the most influential hyperparameters used in machine learning models.
	To test for an effect of the size of the feature vector $\mathbf{z}$ on the classification accuracy, we tried three possible values for $\lvert \mathbf{z} \rvert \in \{100, 1024, 2048\}$.
	
		\begin{table*}[htb!]
		\centering
		\caption{Ablation study -- classification accuracy after 20 tasks.}
		\label{tab:ablation_20}
		\resizebox{\textwidth}{!}{%
			\begin{tabular}{@{}ccc
					>{\columncolor[HTML]{EFEFEF}}r 
					>{\columncolor[HTML]{EFEFEF}}r rr
					>{\columncolor[HTML]{EFEFEF}}r 
					>{\columncolor[HTML]{EFEFEF}}r rr@{}}
				\toprule
				& Task &  & \multicolumn{2}{c}{\cellcolor[HTML]{EFEFEF}$z = 100$} & \multicolumn{2}{c}{$z = 1024$} & \multicolumn{2}{c}{\cellcolor[HTML]{EFEFEF}$z = 2048 (100)$} & \multicolumn{2}{c}{$z = 2048$} \\
				Dataset & known? & Reconstruction & \multicolumn{1}{c}{\cellcolor[HTML]{EFEFEF}Argmax} & \multicolumn{1}{c}{\cellcolor[HTML]{EFEFEF}Random} & \multicolumn{1}{c}{Argmax} & \multicolumn{1}{c}{Random} & \multicolumn{1}{c}{\cellcolor[HTML]{EFEFEF}Argmax} & \multicolumn{1}{c}{\cellcolor[HTML]{EFEFEF}Random} & \multicolumn{1}{c}{Argmax} & \multicolumn{1}{c}{Random} \\ \midrule
				& & \xmark & 96.9 $\pm$ 0.9 & 92.8 $\pm$ 2.1 & 96.5 $\pm$ 0.8 & 98.8 $\pm$ 0.2 & 98.2 $\pm$ 0.3 & 97.0 $\pm$ 0.4 & 98.8 $\pm$ 0.1 & 98.9 $\pm$ 0.2 \\
				& \multirow{-2}{*}{\cmark} & \cmark & \textbf{98.5 $\pm$ 0.2} & 98.4 $\pm$ 0.2 & 98.3 $\pm$ 0.2 & \textbf{99.1 $\pm$ 0.1} & 98.4 $\pm$ 0.3 & \textbf{99.0 $\pm$ 0.1} & 98.8 $\pm$ 0.2 & \textbf{99.0 $\pm$ 0.1} \\ \cmidrule{2-11}
				& & \xmark & 23.3 $\pm$ 5.4 & 61.0 $\pm$ 1.4 & 37.6 $\pm$ 3.2 & 86.2 $\pm$ 0.9 & 39.8 $\pm$ 2.9 & 82.8 $\pm$ 1.1 & 56.6 $\pm$ 3.0 & 87.7 $\pm$ 0.6 \\
				\multirow{-4}{*}{ModelNet40} & \multirow{-2}{*}{\xmark} & \cmark & 67.6 $\pm$ 1.2 & \textbf{79.3 $\pm$ 0.4} & 68.7 $\pm$ 1.9 & \textbf{87.0 $\pm$ 1.2} & 69.1 $\pm$  1.6 & \textbf{86.9 $\pm$ 0.2} & 69.5 $\pm$ 1.1 & \textbf{88.2 $\pm$ 0.2} \\ \midrule
				
				& & \xmark & 96.9 $\pm$ 0.9 & 92.8 $\pm$ 2.1 & 96.5 $\pm$ 0.8 & 98.8 $\pm$ 0.2 & 98.2 $\pm$ 0.3 & 97.0 $\pm$ 0.4 & 98.8 $\pm$ 0.1 & 98.9 $\pm$ 0.2 \\
				& \multirow{-2}{*}{\cmark} & \cmark & \textbf{98.5 $\pm$ 0.2} & 98.4 $\pm$ 0.2 & 98.3 $\pm$ 0.2 & \textbf{99.1 $\pm$ 0.1} & 98.4 $\pm$ 0.3 & \textbf{99.0 $\pm$ 0.1} & 98.8 $\pm$ 0.2 & \textbf{99.0 $\pm$ 0.1} \\ \cmidrule{2-11}
				& & \xmark & 23.3 $\pm$ 5.4 & 61.0 $\pm$ 1.4 & 37.6 $\pm$ 3.2 & 86.2 $\pm$ 0.9 & 39.8 $\pm$ 2.9 & 82.8 $\pm$ 1.1 & 56.6 $\pm$ 3.0 & 87.7 $\pm$ 0.6 \\
				\multirow{-4}{*}{ShapeNet} & \multirow{-2}{*}{\xmark} & \cmark & 67.6 $\pm$ 1.2 & \textbf{79.3 $\pm$ 0.4} & 68.7 $\pm$ 1.9 & \textbf{87.0 $\pm$ 1.2} & 69.1 $\pm$  1.6 & \textbf{86.9 $\pm$ 0.2} & 69.5 $\pm$ 1.1 & \textbf{88.2 $\pm$ 0.2} \\ \bottomrule
			\end{tabular}%
		}
	\end{table*}
	
	PointNet~\cite{qi2017pointnet} has an inherent ability to pick \textit{critical point sets} for a given shape by selecting the points, which feature values passed through the max-pooling layer and create a global feature vector (which we will call \textit{an argmax method}).
	Therefore it feels natural to pick these points for the compressed subset of the point cloud.
	As an added benefit, it would be easy to control the compression level because setting the length of the feature vector to a value $M$ would enforce that \textit{at most} (since points may get picked more than once) $M$ points would be saved to a compressed point cloud.
	The alternative way to pick a subset of points from the point cloud is to draw them randomly, with each point having an equal chance of being picked.
	To ensure a more fair comparison of random and argmax picking (where about 70-80\% of the points were unique), we employ drawing with replacement to keep the number of saved points similar between the two methods.
	
	Usually, for future rehearsal, we select the number of points that is close to the number of features in the latent vector $\mathbf{z}$.
	To check whether these values do not have to be correlated, we conducted an additional experiment to set a feature vector length to 2048 but still subsample 100 points at most. 
	
	In summary, we tested four different point cloud compression strategies for the ablation study -- random and argmax, with or without regularizing the reconstruction module. 
	Each strategy was tested in 6 scenarios on 4 datasets -- MNIST, ModelNet10, ModelNet40 and ShapeNet.
	The study was performed on the distance of 5 and 20 tasks.
	For each scenario and strategy, we picked three sizes of feature vector $\mathbf{z}$ (100, 1024, 2048) that determined how many points were saved from each point cloud.
	We also tested a setting where we used a wide feature vector (of length 2048) and sampled only up to 100 points, denoted in tables as 2048 (100).
	We report the results of all experiments as a mean accuracy of 3 independent runs with a standard deviation value.
	We present the results in \cref{tab:ablation_5} for 5-task and in \cref{tab:ablation_20} for 20-task continual learning.
	
	As a result of the ablation experiments, we can observe that compression by random sampling works better most of the time, especially in the 20-task distance (\cref{tab:ablation_20}).
	Furthermore, adding a reconstruction network often improves overall accuracy in both Task-IL and Class-IL scenarios and the stability of the training process, which we can notice as reduced variability of the final scores.
	This effect is particularly noticeable when using heavier compression and when training more tasks (\cref{tab:ablation_20}).
	We set the length of the feature vector to 2048, and we saved up to 100 points per point cloud due to a significant decrease in storage needed for future rehearsal and a minor loss of total accuracy compared to other compression settings.
	However, for the main experiments, we will resort to sampling without replacing a fixed number of 100 points to reduce the randomness effect on the final results.
	
	\section{Conclusion}
	\label{sec:conclusion}
	In this work, we present a novel approach for continual learning on point cloud data.
	The proposed method is trained in a compressed rehearsal setting. 
	We show that by saving a small subsample of each point cloud in the training set, we can effectively combat catastrophic forgetting in classification tasks.
	Moreover, the presented model works well when the task is known, as well as when it needs to be inferred during classification.
	
    \section*{Acknowledgements}

    The work of T. Trzcinski was supported by the National Centre of Science (Poland) Grant No. 2020/39/B/ST6/01511 as well as the Foundation for Polish Science Grant No. POIR.04.04.00-00-14DE/18-00 co-financed by the European Union under the European Regional Development Fund. The works of M. Zamorski, M. Zieba, and M. Stypulkowski were supported by the National Centre of Science (Poland) Grant No. 2020/37/B/ST6/03463. 

	\bibliographystyle{icml2022}
	\bibliography{bibliography}
\end{document}